\documentclass[runningheads]{llncs}
\usepackage[T1]{fontenc}
\usepackage{graphicx}
\usepackage{cite}
\usepackage{amsmath,amssymb,amsfonts}
\usepackage{algorithmic}
\usepackage{graphicx}
\usepackage{textcomp}
\usepackage{multirow}
\usepackage{subcaption}
\usepackage{marvosym}

\begin{document}
\title{Single-Stream Multi-Feature Fusion with Temporal Robustness for Gait Emotion Recognition}
\titlerunning{Single-Stream Multi-Feature Fusion for Gait Emotion Recognition}
%
\author{Shirong Lyu\inst{1},
Silu Quan\inst{2},
Yixuan Ding\inst{1},\and
Chengpeng Wang\inst{3}\textsuperscript{(\Letter)}}

\authorrunning{S. Lyu et al.}
%
\institute{College of Computer and Information Science, Southwest University, \\
	Chongqing, China \and
	College of Physical Science and Technology, Southwest University, \\
	Chongqing, China \and
	Wisesoft Inc., Chengdu, China\\
\email{wcp12.4@gmail.com}}
\maketitle              
\begin{abstract}
3D skeleton-based gait emotion recognition faces high annotation costs, data scarcity, and poor generalization on heterogeneous data. This paper proposes SV-GCN, a single-stream multi-feature fusion framework with temporal invariance. We introduce intra-frame relative motion features to eliminate frame-rate sensitivity and embed heterogeneous cues at shallow layers, enabling early fusion without multi-stream complexity. For variable-length sequences, we design a global mask-guided valid-frame spatio-temporal graph convolution module, introducing frame-rate insensitivity for the first time in this domain. On the E-Gait dataset, our method achieves performance comparable to state-of-the-art while demonstrating strong generalization across varying sequence lengths and frame rates, offering a viable pathway for pre-training on large-scale skeleton-based action recognition datasets. The code is available at https://github.com/lsr51/SV-GCN.

\end{abstract}
\section{Introduction}
Emotion recognition holds significant application value in human-computer interaction, behavioral analysis, and intelligent surveillance. Research indicates that a substantial portion of affective cues are conveyed through non-verbal signals such as gait and gestures.

Unlike facial expressions, which are susceptible to illumination, distance, and observer effects \cite{b38}, gait provides more reliable affective signals in long-distance scenarios due to its natural and unconscious nature. Advances in 3D pose estimation \cite{b39} have enabled skeleton-based gait emotion recognition. Early studies established kinematic-level correlations between gait and emotion \cite{b26,b27,b28}; for instance, positive emotions often correspond to increased arm swing and faster stride. With deep learning, models like LSTM \cite{b40} and GCN \cite{b41} have been introduced, significantly improving accuracy.

\begin{figure}[t]
\centerline{\includegraphics[width=0.7\textwidth]{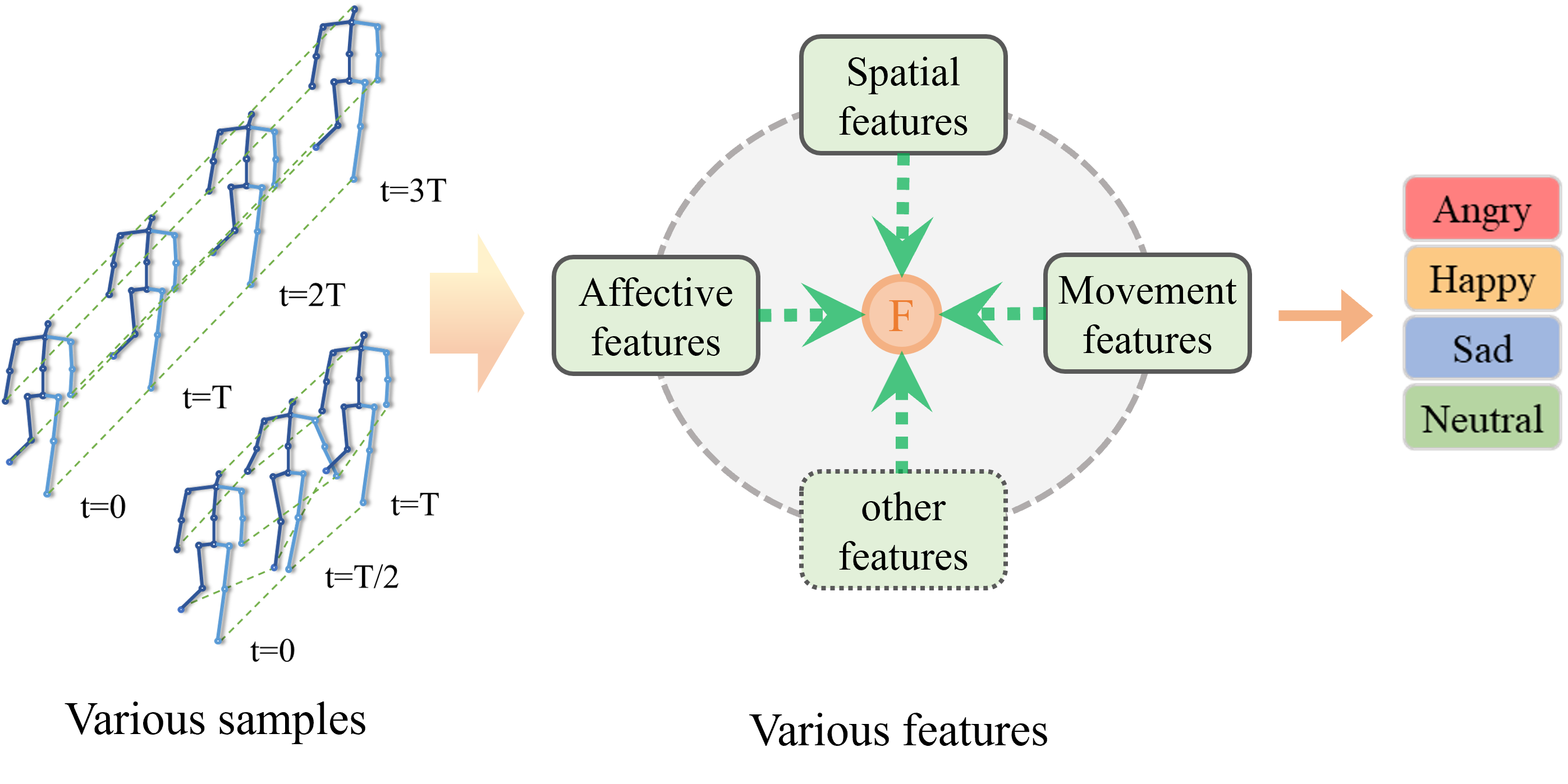}}
\caption{Sample heterogeneity and feature diversity make it difficult to fully utilize information and model gait emotion recognition.} \label{Fig1}
\end{figure}
However, deploying existing gait emotion recognition models for practical applications faces severe challenges, primarily being insufficient generalization. Fig.~\ref{Fig1} illustrates two key factors:

Feature heterogeneity and fusion. STEP \cite{b30} simply concatenates affective features with network-learned features, hindering effective utilization. BPM-GCN \cite{b34} treats affective features as constraints (Affective Constraint) for knowledge distillation, but this indirect guidance is limited, and extending the dual-stream network to multi-stream increases complexity. Affective mapping \cite{b31} imposes only an overall constraint. These methods fail to embed affective features into local joints, preventing shallow-layer feature fusion and weakening fine-grained guidance of affective features on local representations.

Data scarcity and heterogeneous inconsistency. High annotation costs and scarce data are compounded by inconsistencies across datasets (sequence length, frame rate, gait cycle). While some studies employ generative augmentation \cite{b30} or semi-supervised learning \cite{b31} to address sample scarcity, leveraging large-scale skeleton-based action recognition datasets for pre-training remains unexplored. The E-Gait dataset \cite{b30} contains two subsets with different frame rates and lengths, yet existing models \cite{b30,b31,b34} directly ignore these differences, forcibly padding sequences to fixed lengths—causing temporal distortion, feature contamination, and inaccurate batch normalization statistics.

To address the lack of temporal invariance and inefficient multi-feature fusion in existing GCNs, we propose a multi-feature fusion network robust to frame rate and sequence length. First, we introduce a frame-rate insensitivity joint motion feature that captures coordinated body movements, replacing conventional inter-frame absolute velocity \cite{b34}. Second, we embed multiple features (including affective features) into a unified space on a per-joint basis, enabling early fusion via a single-stream ST-GCN \cite{b11} in shallow layers. This avoids parameter explosion and overfitting while learning richer local patterns. Third, we adopt a global masking mechanism (inspired by speech/NLP \cite{b42,b43}) to handle variable-length sequences, ensuring reduction operations use only valid frames and preventing padding contamination. We replace BatchNorm with mask-aware GroupNorm to avoid statistical drift. Meanwhile, we systematically leverage the differences between the two subsets by conducting comparative experiments on separated and combined configurations of the E-Gait dataset. Our main contributions are as follows:

\begin{itemize}
	\item We propose the Shallow Multi-feature Embedding and Fusion module (SMEF), which designs frame-rate insensitivity intra-frame relative motion features and embeds affective features into joints. This achieves the first early fusion of multiple features, overcoming existing methods' limitation of treating affective features only as global constraints.
	\item We propose the Valid Frame Spatio-Temporal Graph Convolutional Network (VF-STGCN), introducing frame-rate insensitivity to gait emotion recognition for the first time. Its global masking mechanism only operates on valid frames, avoiding bias and contamination.
	\item We propose the Single-stream Variable-length Graph Convolutional Network (SV-GCN), which is robust to heterogeneous data with varying lengths and frame rates, and, for the first time, achieves fine‑tuning of a gait emotion model using a model pre‑trained on large‑scale datasets.
\end{itemize}

\section{Related Work}

\subsection{Skeleton-Based Action Recognition}
Early deep learning approaches utilized RNNs and LSTMs \cite{b40} to model skeletal motion \cite{b8}. Subsequently, Graph Convolutional Networks (GCNs) became dominant due to their ability to explicitly model human body structure. Yan et al. \cite{b11} pioneered Spatio-Temporal Graph Convolutional Networks (ST-GCN), laying the foundation for many follow-up works \cite{b14,b15}. In recent years, Transformer-based methods \cite{b17} have also been introduced to capture global dependencies. Widely used datasets include NTU RGB+D \cite{b20} and Kinetics \cite{b21}, which serve as important benchmarks for evaluating model generalization. For a comprehensive review, refer to \cite{b22}.

\subsection{Gait Emotion Recognition}
Early research validated the feasibility of gait-based emotion recognition using handcrafted features combined with classical methods such as SVM \cite{b26}. With the advancement of deep learning, Randhavane et al. \cite{b29} fused handcrafted features with LSTMs for emotion recognition. Following the introduction of ST-GCN \cite{b11}, a series of GCN-based methods emerged. Bhattacharya et al. \cite{b30} created the E-Gait dataset and proposed STEP; they later proposed a semi-supervised method \cite{b31}. Other representative works include BPM-GCN \cite{b34} and MSA-GCN \cite{b35}. Due to dataset scarcity and overfitting concerns, Transformer-based methods remain largely unexplored in this domain.

\section{Method}
\begin{figure}[t]
	\centering
	\includegraphics[width=0.98\textwidth]{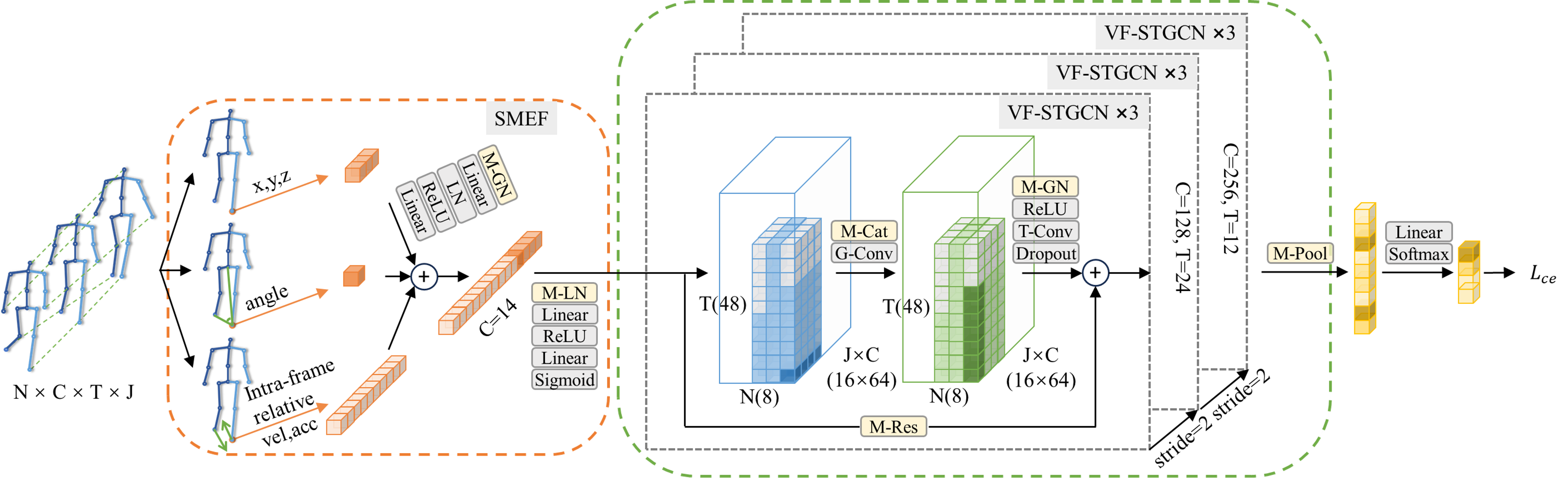}
	\caption{Overview of the proposed SV-GCN framework. The SMEF module (red) performs multi-feature embedding and fusion at joint level; the VF-STGCN module (green) handles variable-length sequences guided by a global valid frame mask. $N$, $C$, $T$, $J$ denote batch size, channels, max frames, and number of joints, respectively. Yellow boxes indicate mask-aware modules.}
	\label{Fig2}
\end{figure}

This section presents SV-GCN, a single-stream multi-feature fusion framework with a valid frame mechanism for robust gait emotion recognition. As illustrated in Fig.~\ref{Fig2}, the framework consists of two key components: the Single-stream Multi-feature Embedding and Fusion (SMEF) module, which integrates heterogeneous features at the joint level with adaptive weighting, and the Valid Frame Spatio-Temporal Graph Convolutional (VF-STGCN) module, which employs a global masking strategy to handle variable-length sequences while preserving statistical stability. Together, these designs enhance the model's robustness to temporal variations and multi-source heterogeneity.

\subsection{Single-stream Multi-feature Embedding and Fusion Module}
This module (SMEF) employs a single-stream network to perform joint-based local representation and fusion of different types of features at the input stage.

The three-dimensional joint coordinates $f_{\text{Coord}}(x, y, z)$ serve as the primary foundational pose features.

\begin{figure}[t]
	\centerline{\includegraphics[width=0.53\textwidth]{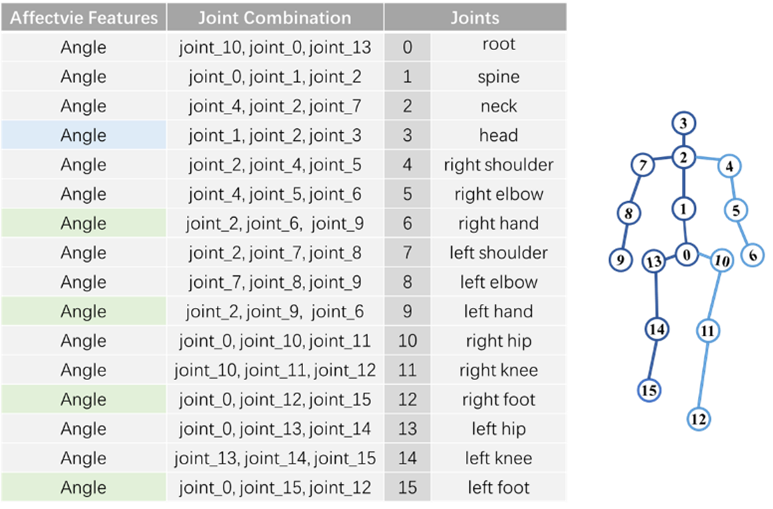}}
	\caption{Affective feature embeddings correspond to joint points}
	\label{Fig3}
\end{figure}

Affective features describe geometric relationships among joints, such as angles, distances, and areas. These are typically defined by 2–3 joints. Although they are not attributes of a single node, they have a clear dominant node in spatial relationships (except for area features). For example, angle features are dominated by the middle joint. Drawing on the affective feature sets defined in papers \cite{b31,b34}, we associate each joint with the most relevant affective features, with the mapping relationship shown in Fig.~\ref{Fig3}. For limb joints, the angle feature is assigned to the middle joint; for endpoints (head, hands, feet), we design specific angle features as indicated in blue and green in the figure.

For three adjacent joints on a limb connection, the included angle is calculated as the affective feature of the middle joint. Given three-dimensional joint coordinates $J_1, J_2, J_3 \in \mathbb{R}^3$ (Fig.~\ref{Fig3}, second column), the angle feature $f_\theta$ is expressed by \eqref{eq1}.
\begin{equation}
	\begin{aligned}
		u &= \frac{J_1 - J_2}{\|J_1 - J_2\|_2}, \quad
		v = \frac{J_3 - J_2}{\|J_3 - J_2\|_2}, \\
		f_\theta &= \arccos\!\left( \min\!\left\{ \max\{u \cdot v, -1\}, 1 \right\} \right) \in [0, \pi]
	\end{aligned}
	\label{eq1}
\end{equation}

To capture dynamic information and enhance the model's robustness to data with varying frame rates, we introduce features based on relative motion rather than using absolute velocities that depend on time intervals. Let ${f}_{\Delta t}^{(j)} \in \mathbb{R}^{10}$ denote the motion feature [linear vel. (4D: $x,y,z$, mag), linear acc. (4D: $x,y,z$, mag), angular vel. (1D = $f_\theta$), angular acc. (1D)] of node $j$ between adjacent frames. By calculating the relative motion state of each joint with respect to the root node within the same frame, we eliminate the direct influence of temporal variables, thereby providing a time-scale-invariant description of coordinated body-part movements. The relative motion feature ${f}_{\text{rel}}^{(j)} \in \mathbb{R}^{10}$ of node $j$ with respect to the root node $0$ is computed as in \eqref{eq2} and \eqref{eq3}, where $\epsilon = 10^{-8}$ prevents division by zero, $\odot$ is element-wise multiplication, the logarithm stabilizes gradients, and $\sigma$ represents the relative direction.
\begin{equation}
	\begin{aligned}
		{f}_{\mathrm{rel}}^{(j)} = {\sigma} \odot \ln\left( \frac{\|{f}_{\Delta t}^{(j)} + \epsilon\|}{\|{f}_{\Delta t}^{(0)} + \epsilon\|} \right)
		\label{eq2} \\
	\end{aligned}
\end{equation}
\begin{equation}
	\begin{aligned}
		{\sigma} \triangleq \operatorname{sign}({f}_{\Delta t}^{(j)}) \odot \operatorname{sign}({f}_{\Delta t}^{(0)}) \in \{-1, +1\}^{10}
		\label{eq3}
	\end{aligned}
\end{equation}

The above pose features, motion features, and affective features (totaling $14$ dimensions) are embedded and concatenated along the channel dimension within the network to form the initial fused feature ${X} \in \mathbb{R}^{N \times C \times T \times J}$, where $N, C, T, J$ denote batch size, number of channels, time steps, and number of joints, respectively.

We introduce a channel attention mechanism to allow the network to adaptively learn the importance of different feature channels. Considering that the features are located in the shallow layers of the network and their numerical scales may vary significantly, we first apply LayerNorm and linear projection to the features, as shown in \eqref{eq4}, where $\operatorname{Proj}(\cdot)$ denotes a linear transformation layer combined with a ReLU activation function.
\begin{multline}
	{X}_{\text{embed}} = \bigl[ 
	\operatorname{Proj}\!\left( \operatorname{LN}\!\left( {X}_{(:, [\text{Coord}], :, :)} \right) \right), \\
	\operatorname{Proj}\!\left( \operatorname{LN}\!\left( {X}_{(:, [f_{rel}[0:7]], :, :)} \right) \right), \\
	\operatorname{Proj}\!\left( \operatorname{LN}\!\left( {X}_{(:, [f_{\theta}, f_{rel}[8:9]], :, :)} \right) \right)
	\bigr]
	\label{eq4}
\end{multline}

Apply LayerNorm to ${X}_{\text{embed}}$, then average pool over $T$ and $J$, and generate channel-wise attention weights $w_c^i$ using a two-layer MLP with sigmoid activation. By employing the mask \eqref{eq7}, only valid frames are considered. The weighted feature output is given by \eqref{eq6}.
\begin{equation}
	\begin{aligned}
		w_c^{i} = \sigma\left(\mathrm{Linear}_2\left(\mathrm{ReLU}\left(\mathrm{Linear}_1\left(\frac{1}{J L_i} \sum_{t=0}^{L_i-1} \sum_{j=0}^{J-1} X_{i, :, t, j}\right)\right)\right)\right)  \label{eq5}
	\end{aligned}
\end{equation}
\begin{equation}
	\begin{aligned}
		X_{\mathrm{attn}} = X_{i,c,t,j}^{\mathrm{Embed}} \cdot w_c^i \cdot M_{i,t} \label{eq6}
	\end{aligned}
\end{equation}

\subsection{Adaptive Alignment Valid Frame ST-GCN}
To effectively handle variable-length gait sequences while avoiding overfitting, we introduce a global masking mechanism based on ST-GCN and propose the Valid Frame ST-GCN module (VF-STGCN). This design ensures that all temporal aggregation operations are applied only to valid data regions. A binary mask ${M} \in \{0, 1\}^{N \times T}$ indicates the valid length $L_i$ of sample $i$, broadcast to ${M}_{\text{4d}} \in \{0, 1\}^{N \times C \times T \times J}$ with ${M}_{\text{4d}} = M_{n,t} \; (\forall_{c,j})$. The valid computation domain $\Omega$ is defined as in \eqref{eq7}. All temporal reduction operations are constrained to $\Omega$.
\begin{equation}
	\begin{aligned}
		M_{i,t} = 
		\begin{cases}
			1, & t < L_i \\
			0, & \text{otherwise}
		\end{cases}, \qquad
		\Omega = \{(n,c,t,j) \mid M_{n,c,t,j} = 1\}
		\label{eq7}
	\end{aligned}
\end{equation}

However, in the presence of masking, the statistics of BatchNorm fluctuate with the valid count, leading to training instability. Therefore, we employ Masked GroupNorm (M-GN) as a replacement. M-GN divides the channels into $G$ groups and independently computes the normalization statistics based solely on the valid frames within the current sample and the current group. For the $g$-th group, the mean $\mu_g$ and variance $\sigma_g^2$ are calculated as shown in \eqref{eq9}, where $\mathrm{valid\_count}_{g} = \sum_{n,c,t,j} M_{n,g,c,t,j}^{(g)}$. This eliminates the influence of inter-sample length variation on the statistical distribution.
\begin{equation}
	\begin{aligned}
		\mu_g = \frac{\displaystyle\sum_{n,c,t,j} X_{n,g,c,t,j}^{(g)} \cdot M_{n,g,c,t,j}^{(g)}}{\mathrm{valid\_count}_{g}}, \quad
		\sigma_g^2 = \frac{\displaystyle\sum_{n,c,t,j} \left( X_{n,g,c,t,j}^{(g)} - \mu_g \right)^2 \cdot M_{n,g,c,t,j}^{(g)}}{\mathrm{valid\_count}_{g}}
		\label{eq9}
	\end{aligned}
\end{equation}

To prevent invalid frames from propagating through the adjacency matrix $A$, valid frames are extracted and concatenated before graph convolution. The valid features of all samples are concatenated into a compact continuous tensor, on which standard graph convolution is performed, and the results are redistributed back according to the original sample order, as described in \eqref{eq11}. Here, $T_{\text{total}} = \sum_{i=1}^{N} L_i$ denotes the total length of all valid frames in the batch.
\begin{equation}
	\begin{aligned}
		&{X}_i^{\text{valid}} = {X}_i[:, :L_i, :] \in \mathbb{R}^{C \times L_i \times J} \\
		&{X}^{\text{compact}} = \operatorname{Concat}\left( {X}_1^{\text{valid}}, {X}_2^{\text{valid}} \dots, {X}_N^{\text{valid}} \right) \in \mathbb{R}^{C \times T_{\text{total}} \times J} \\
		&{Y}^{\text{compact}}, {A}' = {G}\left( {X}^{\text{compact}}, {A} \right) \\
		&{Y}_i[:, :L_i, :] = {Y}^{\text{compact}}[:, t_{i-1}:t_i, :]
		\label{eq11}
	\end{aligned}
\end{equation}

Furthermore, for downsampling operations with stride $s > 1$ (used in TCN and residual connections), the valid frame length is updated to $L_i' = \big\lceil L_i / s \big\rceil$, and the temporal mask is correspondingly adjusted via pooling.
\begin{equation}
	\begin{aligned}	
		Y_{n,c} = \frac{\displaystyle\sum_{t=1}^{T} \sum_{j=1}^{J} M_{n,t} \cdot X_{n,c,t,j}}{\max(J \cdot L_n, 1)}
		\label{eq15} \\
	\end{aligned}
\end{equation}
\begin{equation}
	\begin{aligned}	
		{Z}_n = \operatorname{FC}\left( {Y}_n \right)
		\label{eq16} \\
	\end{aligned}
\end{equation}
\begin{equation}
	\begin{aligned}	
		\mathcal{L}_{\mathrm{CE}}^{(n)} = -\log\left( \frac{\exp(z_{n,y_n})}{\sum_{c=1}^{C} \exp(z_{n,c})} \right)
		\label{eq17}
	\end{aligned}
\end{equation}

The network backbone consists of three stacked groups of VF-STGCN modules, each containing three blocks with channel dimensions of $64$, $128$, and $256$, respectively. Downsampling between groups is performed via temporal convolution with stride $s = 2$; the temporal kernel sizes are set to $11$, $7$, and $3$ successively to accommodate short‑sequence samples. The final features are aggregated by masked global average pooling, as shown in \eqref{eq15}, where $L_n = \sum_{t=1}^{T} M_{n,t}$ denotes the number of valid frames of sample $n$. The pooled features are then mapped through a fully‑connected layer and optimized using the standard cross‑entropy loss, as described in \eqref{eq16} and \eqref{eq17}.

\section{Experiments}
\subsection{Datasets}
The Emotion-Gait (E-Gait) dataset \cite{b30} is the only popular publicly available dataset, comprising 2,177 real and 1,000 synthetic 3D skeletal samples. Each sample consists of 16 joints with 3-dimensional coordinates. Among the real gaits, 1,835 samples are taken from the Edinburgh Locomotion MOCAP Database \cite{b2}, where each running or walking gait contains 240 frames over 4 seconds. The remaining 342 samples are walking data collected by the authors, with frame counts varying between 18 and 75. We leverage the inherent differences between the two subsets (EG-1835 and EG-342) to evaluate generalization under heterogeneous source conditions. The real gaits were labeled with four emotion classes (happy, sad, angry, and neutral) by the same domain experts. Following \cite{b34}, for the 1,835 samples, we applied fixed-stride (t=5) downsampling on the 240-frame sequences to obtain samples of length T=48. For the 342 samples, we padded all sequences to a uniform length of 75 frames by repeating the last frame and truncating to the first 48 frames, aligning the temporal dimension across subsets. The data were split into training and test sets in a 9:1 ratio.

\subsection{Implementation Details}
The default experimental settings include an initial learning rate of 0.008, weight decay of $1 \times 10^{-4}$, a batch size of 12, and a total of 200 training epochs. We employ the Adam optimizer. To mitigate gradient instability during the initial training phase, we introduce a learning rate warm-up mechanism: over the first 7 epochs, the learning rate linearly increases from 0.0008 to 0.008. After the warm-up phase, an exponential decay strategy is applied with a decay factor of $\gamma = 0.99$. We randomly apply data augmentation strategies such as temporal scaling, temporal shifting, Gaussian noise-based coordinate jittering, and spatial rotation with certain probabilities to enhance sample diversity and prevent overfitting. All experiments are conducted on a single NVIDIA GeForce 3090 GPU.

\begin{table}[t]
	\centering
	
	\begin{minipage}[c]{0.49\textwidth}
		\centering
		\captionof{table}{Accuracy Comparison of Feature Embeddings}
		\renewcommand{\arraystretch}{1.0}
		\begin{tabular}{c|c c c c|c}
			\hline
			\multirow{2}{*} & {Joint} & {Affective} & {Inter} & \multirow{2}{*}{{CA}} & \multirow{2}{*}{{Acc(\%)}} \\
			& {Coords.} & {Feat.} & {Vel.} & & \\
			\hline
			1 & $\checkmark$ & & & & 86.96 \\
			2 & $\checkmark$ & $\checkmark$ & & & 87.86 \\
			3 & $\checkmark$ & $\checkmark$ & $\checkmark$ & & 90.76 \\
			4 & $\checkmark$ & $\checkmark$ & $\checkmark$ & $\checkmark$ & 92.39 \\
			\hline
		\end{tabular}
		\label{tab1}
	\end{minipage}
	\hfill
	\begin{minipage}[c]{0.49\textwidth}
		\centering
		\captionof{table}{Accuracy Comparisons between Fixed-length and Variable-length Datasets}
		\renewcommand{\arraystretch}{1.0}
		\begin{tabular}{c|c|c c c c|c}
			\hline
			\multirow{2}{*}{\shortstack{Dataset}} & \multirow{2}{*}{} & \multirow{2}{*}{Base} & \multirow{2}{*}{VF} & \multirow{2}{*}{\shortstack{Intra\\Vel.}} & \multirow{2}{*}{Kernel} & \multirow{2}{*}{Acc(\%)} \\
			& & & & & & \\
			\hline
			\multirow{4}{*}{EG-1835} & 1 & $\checkmark$ & & & & 90.76\\
			& 2 & $\checkmark$ & $\checkmark$ & & & 88.59\\
			& 3 & $\checkmark$ & $\checkmark$ & $\checkmark$ & & 89.13\\
			& 4 & $\checkmark$ & $\checkmark$ & $\checkmark$ & $\checkmark$ & 89.67\\
			\hline
			\multirow{4}{*}{\shortstack{EG-1835\\+EG-342}} & 5 & $\checkmark$ & & & & 88.07\\
			& 6 & $\checkmark$ & $\checkmark$ & & & 88.53\\
			& 7 & $\checkmark$ & $\checkmark$ & $\checkmark$ & & 89.00\\
			& 8 & $\checkmark$ & $\checkmark$ & $\checkmark$ & $\checkmark$ & 89.45\\
			\hline
		\end{tabular}
		\label{tab2}
	\end{minipage}
\end{table}

\subsection{Ablation Studies}
We adopt the code from \cite{b30} as baseline and report accuracy averaged over three runs.

\textbf{Analysis of multi-feature embedding}. Table~\ref{tab1} evaluates the effectiveness of the proposed multi-feature embedding module on the EG-1835 subset, where all samples have uniform length and frame rate. Rows 1–3 progressively incorporate joint coordinates, affective features, and movement features, showing clear accuracy gains. Row 4 further introduces channel-wise attention, leading to the best performance.

\textbf{Analysis of the impact of sequence-length and frame-rate consistency on model training.} Table~\ref{tab2} compares two data configurations: EG-1835 (uniform) and EG-1835+EG-342 (variable-length). The label-to-sample counts in EG-1835 are (0:1048, 1:454, 2:254, 3:79), while EG-342 has (0:112, 1:33, 2:78, 3:119). The baseline uses joint coordinates, affective features, and inter-frame absolute velocity, with BatchNorm and a fixed kernel size of 9. Rows 2–4 and 6–8 progressively replace components with the proposed VF-STGCN module, intra-frame velocity features, and kernel sizes of 9, 7, 5. After adopting VF-STGCN, BatchNorm is replaced with GroupNorm (32 channels per group), with dropout 0.2.

\begin{figure}[t]
	\centering
	\begin{subfigure}[t]{0.37\textwidth}
		\centering
		\includegraphics[width=\linewidth]{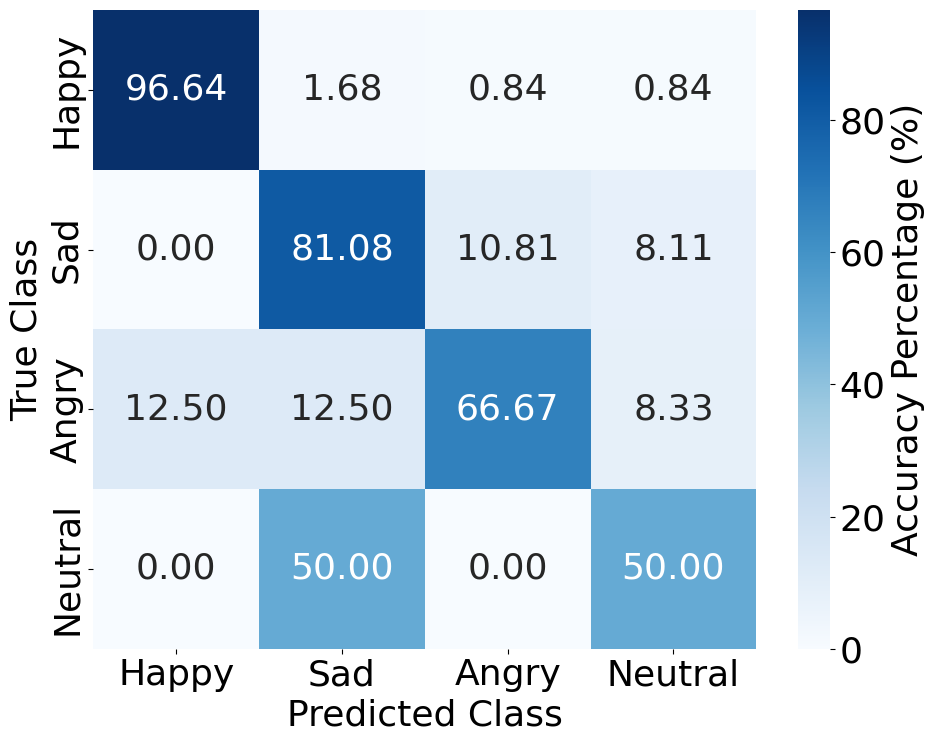}
		\caption{EG-1835}
		\label{Fig4.1.a}
	\end{subfigure}
	\hspace{0.03\textwidth}
	\begin{subfigure}[t]{0.37\textwidth}
		\centering
		\includegraphics[width=\linewidth]{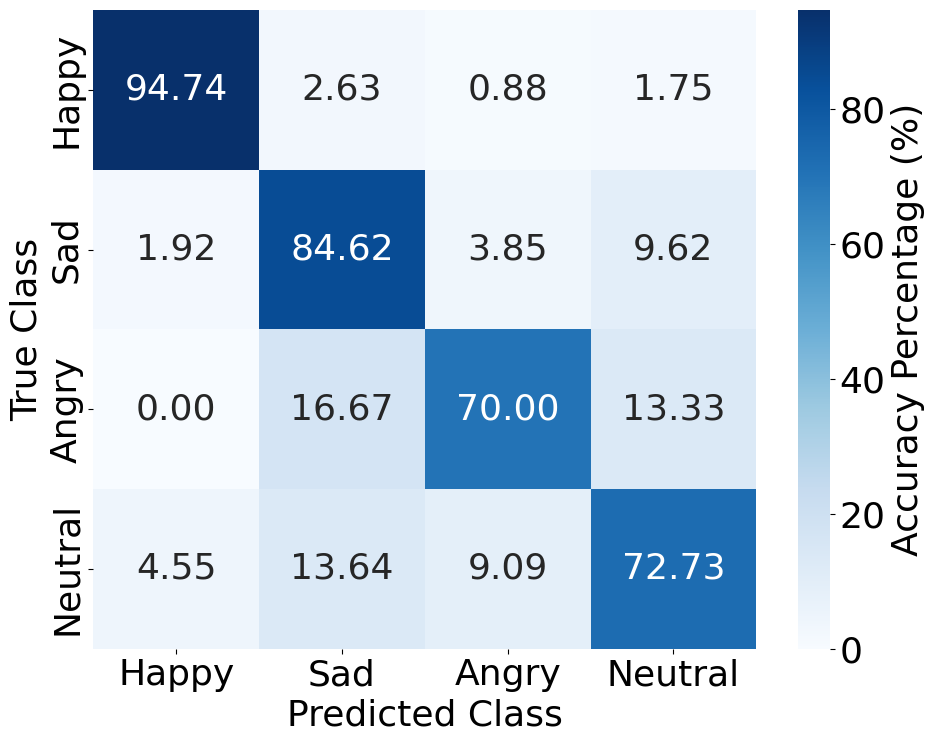}
		\caption{EG-1835 + EG-342}
		\label{Fig4.1.b}
	\end{subfigure}
	\caption{Comparison of Confusion Matrices for Models Trained on Different Datasets}
	\label{Fig4.1}
\end{figure}

 As shown in Table~\ref{tab2}, adding the variable-length EG-342 subset increases training difficulty and lowers overall accuracy (row 1 vs. row 5). However, in this challenging setting, the proposed modifications consistently improve accuracy (rows 6–8 vs. row 5), demonstrating their effectiveness. Fig.~\ref{Fig4.1} shows confusion matrices for the configurations in rows 4 and 8. The model trained only on EG-1835 achieves high overall accuracy but performs poorly on the minority Neutral class (Fig.~\ref{Fig4.1.a}), while incorporating EG-342 yields more balanced predictions across all classes (Fig.~\ref{Fig4.1.b}).

\subsection{Comparison with SOTAs}
Based on the configuration in Table~\ref{tab2} (row 8), we convert the large-scale skeleton-based action recognition dataset NTU RGB+D~\cite{b20} from 25 joints to 16 joints, select 43,515 samples for pretraining, our method achieves competitive accuracy against state-of-the-art approaches, as summarized in Table~\ref{tab3}. Compared to the baseline STEP~\cite{b30}, our method yields a significant improvement. Notably, while dual-stream methods like 2s-AGCN~\cite{b14}, TNTC~\cite{a4}, and BPM-GCN~\cite{b34} rely on higher complexity or interaction modules, our simple single-stream network with early multi-feature fusion attains comparable performance, demonstrating the effectiveness of the proposed design.

\subsection{Visualization}
\begin{figure}[t]
	\centering
	\begin{minipage}[c]{0.63\textwidth}
		\centering
		\includegraphics[width=\linewidth]{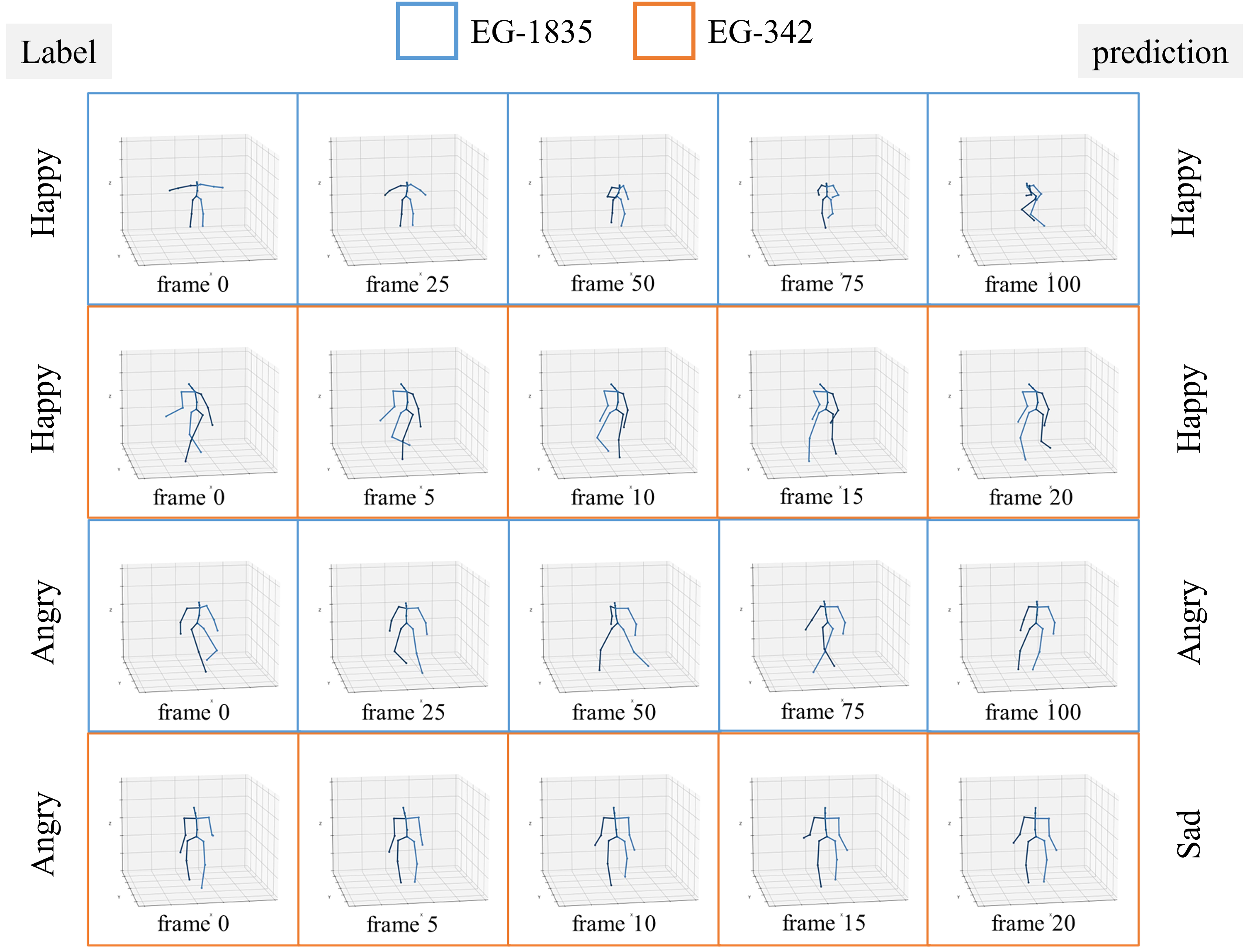}
		\caption{Gait sequence alignment from heterogeneous data (only 5 frames are shown). Frame \# indicates the sequence number of the corresponding frame in its original sample sequence.}
		\label{Fig4.3}
	\end{minipage}
	\hfill
	\begin{minipage}[c]{0.35\textwidth}
		\centering
		\small
		\captionof{table}{Accuracy Comparisons of Different Methods on the Emotion-Gait Dataset}
		\renewcommand{\arraystretch}{1.0}
		\setlength{\tabcolsep}{2pt}
		\begin{tabular}{c|c}
			\hline
			Methods & Acc (\%) \\
			\hline
			LSTM\cite{a1} & 75.10 \\
			G-GCSN\cite{a2} & 81.50 \\
			ProxEmo\cite{b32} & 82.40 \\
			STEP\cite{b30} & 83.15 \\
			2s-AGCN\cite{b14} & 84.40 \\
			TNTC\cite{a4} & 85.97 \\
			ST-Gait++\cite{a5} & 87.50 \\
			BPM-GCN\cite{b34} & 90.37 \\
			\textbf{SV-GCN(Ours)} & 89.91 \\
			\hline
		\end{tabular}
		\label{tab3}
	\end{minipage}
\end{figure}

The two subsets of the E-Gait dataset exhibit significant differences in gait patterns. EG-1835 contains diverse motion modes (walking, jogging, running) and spans about 4 seconds with four gait cycles, while EG-342 was collected under the specific scenario of “walk while thinking of the four different emotions,” lasting only about 1 second with one gait cycle. Using the method from \cite{b6}, we rendered the joint data into dynamic 3D videos (Fig.~\ref{Fig4.3}). During preprocessing, EG-1835 was downsampled every 5 frames, whereas EG-342 was not. As shown in the figure, EG-1835 exhibits more pronounced pose variations over time. The limited valid frames and low inter-frame variability in EG-342 weaken classification performance, making the model especially prone to confusing ‘angry’ with ‘sad’.

\section{Conclusion}
This paper addresses key limitations in gait emotion recognition—inefficient multi-feature fusion, frame-rate sensitivity, and padding interference—by proposing a unified framework that enables effective training on heterogeneous data, consisting of a shallow multi-feature embedding and fusion module (SMEF) for joint-level affective feature integration and a valid-frame spatio-temporal graph convolution module (VF-STGCN) with global masking for robust variable-length sequence handling. Experiments on the E-Gait dataset demonstrate that the proposed SV-GCN achieves competitive performance, and its robustness on heterogeneous subsets suggests strong potential for transfer learning from large-scale skeleton-based action recognition datasets, offering a viable pathway to alleviate the poor generalization of gait emotion models.

%
%
%

\begin{thebibliography}{8}
	\bibitem{b2} I. Habibie, D. Holden, J. Schwarz, J. Yearsley, and T. Komura, A recurrent variational autoencoder for human motion synthesis, In Proceedings of the British Machine Vision Conference (BMVC) (2017).
	\bibitem{b6} W. Zhu, X. Ma, Z. Liu, L. Liu, W. Wu, and Y. Wang, MotionBERT: A unified perspective on learning human motion representations, In Proceedings of the IEEE/CVF Conference on Computer Vision and Pattern Recognition (CVPR) (2023).
	\bibitem{b8} J. Liu, A. Shahroudy, D. Xu, and G. Wang, Hierarchical recurrent neural network for skeleton based action recognition, In Proceedings of the IEEE/CVF Conference on Computer Vision and Pattern Recognition (CVPR), pp. 1110–1118 (2015).
	\bibitem{b11} S. Yan, Y. Xiong, and D. Lin, Spatial temporal graph convolutional networks for skeleton-based action recognition, In Proceedings of the AAAI Conference on Artificial Intelligence (AAAI) (2018).
	\bibitem{b14} L. Shi, Y. Zhang, J. Cheng, and H. Lu, Two-stream adaptive graph convolutional networks for skeleton-based action recognition, In Proceedings of the IEEE/CVF Conference on Computer Vision and Pattern Recognition (CVPR), pp. 12 026–12 035 (2019).
	\bibitem{b15} Z. Liu, H. Zhang, Z. Chen, Z. Wang, and W. Ouyang, Disentangling and unifying graph convolutions for skeleton-based action recognition, In Proceedings of the IEEE/CVF Conference on Computer Vision and Pattern Recognition (CVPR), pp. 143–152 (2020).
	\bibitem{b17} C. Zheng, S. Zhu, M. Mendieta, T. Yang, C. Chen, and Z. Ding, 3D human pose estimation with spatial and temporal transformers, In Proceedings of the IEEE/CVF International Conference on Computer Vision, pp. 11 656–11 665 (2021).
	\bibitem{b20} A. Shahroudy, J. Liu, T.-T. Ng, and G. Wang, NTU RGB+D: A large scale dataset for 3D human activity analysis, in Proc. In Proceedings of the IEEE/CVF Conference on Computer Vision and Pattern Recognition (CVPR), pp. 1010–1019 (2016).
	\bibitem{b21} J. Carreira and A. Zisserman, Quo vadis, action recognition? A new model and the Kinetics dataset, In Proceedings of the IEEE/CVF Conference on Computer Vision and Pattern Recognition (CVPR), pp. 6299–6308 (2017).
	\bibitem{b22} M. Liu, H. Liu, Q. Hu, B. Ren, J. Yuan, J. Lin, and J. Wen, 3D skeleton-based action recognition: A review, arXiv:2506.00915 (2025).
	\bibitem{b26} G. Venture, H. Kadone, T. Zhang, J. Grèzes, A. Berthoz, and H. Hicheur, Recognizing emotions conveyed by human gait, International Journal of Social Robotics, pp. 621–632 (2014).
	\bibitem{b27} B. Li, C. Zhu, S. Li, and T. Zhu, Identifying emotions from non-contact gaits information based on Microsoft Kinects, IEEE Transactions on Affective Computing, pp. 585–591 (2016).
	\bibitem{b28} M. Chiu, J. Shu, and P. Hui, Emotion recognition through gait on mobile devices, In: IEEE PerCom Workshops (2018).
	\bibitem{b29} T. Randhavane, U. Bhattacharya, P. Kabra, K. Kapsaskis, K. Gray, D. Manocha, and A. Bera, Learning gait emotions using affective and deep features, In: Proceedings of the 15th ACM SIGGRAPH Conference on Motion, Interaction and Games, pp. 1–10 (2022).
	\bibitem{b30} U. Bhattacharya, T. Mittal, R. Chandra, T. Randhavane, A. Bera, and D. Manocha, STEP: Spatial temporal graph convolutional networks for emotion perception from gaits, In Proceedings of the AAAI Conference on Artificial Intelligence (AAAI), pp. 1342–1350 (2020).
	\bibitem{b31} U. Bhattacharya, C. Roncal, T. Mittal, R. Chandra, K. Kapsaskis, K. Gray, A. Bera, and D. Manocha, Take an emotion walk: Perceiving emotions from gaits using hierarchical attention pooling and affective mapping, In Proceedings of the European Conference on Computer Vision (ECCV), pp. 145–163 (2020).
	\bibitem{b32} V. Narayanan, B. M. Manoghar, V. S. Dorbala, D. Manocha, and A. Bera, ProxEmo: Gait-based emotion learning and multi-view proxemic fusion for socially-aware robot navigation, In Proceedings of the IEEE/RSJ International Conference on Intelligent Robots and Systems (IROS) (2020).
	\bibitem{b34} Y. Zhai, G. Jia, Y.-K. Lai, J. Zhang, J. Yang, and D. Tao, Looking into gait for perceiving emotions via bilateral posture and movement graph convolutional networks, IEEE Transactions on Affective Computing (2024).
	\bibitem{b35} Y. Yin, L. Jing, F. Huang, G. Yang, and Z. Wang, MSA-GCN: Multiscale adaptive graph convolution network for gait emotion recognition, Pattern Recognition (2024).
	\bibitem{b38} J.-M. Fernández-Dols and M.-A. Ruiz-Belda, Expression of emotion versus expressions of emotions, In: Everyday Conceptions of Emotion, pp. 505–522 (1995).
	\bibitem{b39} R. B. Neupane, K. Li, and T. F. Boka, A survey on deep 3D human pose estimation, Artificial Intelligence Review, vol. 58, no. 1 (2025).
	\bibitem{b40} S. Hochreiter and J. Schmidhuber, Long short-term memory, Neural Comput., vol. 9, no. 8, pp. 1735–1780 (1997).
	\bibitem{b41} T. N. Kipf and M. Welling, Semi-supervised classification with graph convolutional networks, In Proceedings of the International Conference on Learning Representations (ICLR) (2017).
	\bibitem{b42} A. Vaswani et al., Attention is all you need, In Advances in Neural Information Processing Systems (NeurIPS), pp. 5998–6008 (2017).
	\bibitem{b43} D. Bahdanau, K. Cho, and Y. Bengio, Neural machine translation by jointly learning to align and translate, In Proceedings of the International Conference on Learning Representations (ICLR) (2015).
	\bibitem{a1} T. Randhavane, U. Bhattacharya, K. Kapsaskis, K. Gray, A. Bera, and D. Manocha, Learning perceived emotion using affective and deep features for mental health applications, In: IEEE International Symposium on Mixed and Augmented Reality Adjunct (ISMAR-Adjunct) (2019).
	\bibitem{a2} Y. Zhuang, L. Lin, R. Tong, J. Liu, Y. Iwamot, Y.-W. Chen, G-gcsn: Global graph convolution shrinkage network for emotion perception from gait, In Proceedings of the Asian Conference on Computer Vision (ACCV) (2020).
	\bibitem{a4} C. Hu, W. Sheng, B. Dong, and X. Li, Tntc: Two-stream network with transformer-based complementarity for gait-based emotion recognition, In Proceedings of the IEEE International Conference on Acoustics, Speech and Signal Processing (ICASSP) (2022).
	\bibitem{a5} L. M. Lima, W. D. L. Costa, E. T. Martinez, V. Teichrieb, ST-Gait++: Leveraging spatio-temporal convolutions for gait-based emotion recognition on videos, In the IEEE/CVF Conference on Computer Vision and Pattern Recognition Workshops (CVPRW) (2024).
\end{thebibliography}
%

\begin{figure}
	\centering
	\includegraphics[width=0.4\linewidth]{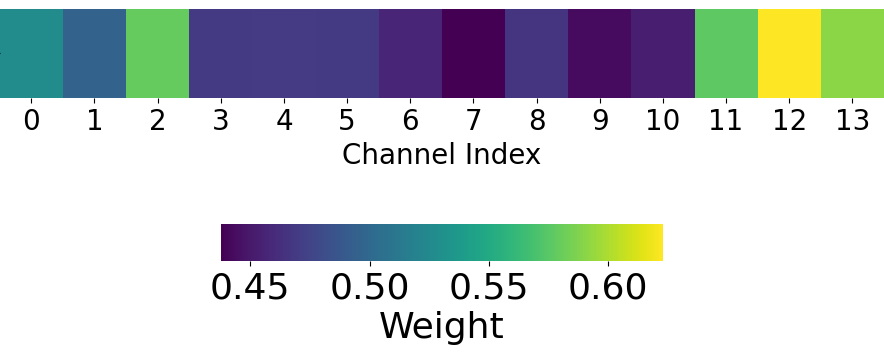}
	\caption{Multi-feature embedding channel attention}
	\label{Fig4.2}
\end{figure}

\section{Supplementary Material}
We performed a visual analysis of the attention weight vector in Eq. (5) of the main text, and the results are shown in Fig.~\ref{Fig4.2}. Channels 0-2 correspond to the three-dimensional (x, y, z) joint-coordinate features; the bright yellow regions indicate the importance of these basic pose features. Channels 11-13 correspond to the affective features we introduced (which are mainly composed of joint angles) as well as angular-velocity and angular-acceleration features. The prominent yellow highlighting indicates that joint rotation has a strong representational effect on affect. Channels 3-10 represent the linear-velocity and linear-acceleration features of the joints. Although their weights appear as relatively dark blue in the visualization, the values remain around 0.4, confirming the effectiveness of motion features in the overall feature composition. Furthermore, considering that replacing BatchNorm with GroupNorm reduces the number of valid samples available for computing statistics, we did not compute channel‑wise attention independently for each joint, so as to avoid introducing noise due to insufficient sample size.

\end{document}